\documentclass[11pt]{article}

\usepackage[preprint]{acl}
\usepackage{float}
\usepackage{times}
\usepackage{latexsym}
\usepackage{listings}
\usepackage{amsmath}
\usepackage[most]{tcolorbox}
\tcbuselibrary{listings,breakable,skins}
\usepackage{makecell} 
\usepackage[T1]{fontenc}

\usepackage[utf8]{inputenc}

\usepackage{microtype}

\usepackage{inconsolata}

\usepackage{graphicx}

\title{Think Twice Before You Write - an Entropy-based Decoding Strategy to Enhance LLM Reasoning}
 
\author{
  Jiashu He\textsuperscript{1,2}\thanks{These authors contributed equally.}\thanks{Work done during internship at Oracle AI.} \quad
  Meizhu Liu\textsuperscript{1} \footnotemark[1]\quad
  Olaitan P Olaleye\textsuperscript{1} \footnotemark[1]\quad
  Amit Agarwal\textsuperscript{1} \\
  \textbf{M. Avendi\textsuperscript{1}} \quad
  \textbf{Yassi Abbasi\textsuperscript{1}} \quad
  \textbf{Matthew Rowe\textsuperscript{1}} \quad
  \textbf{Hitesh Laxmichand Patel\textsuperscript{1}} \\
  \textbf{Paul Li\textsuperscript{1}} \quad
  \textbf{Tao Sheng\textsuperscript{1}} \quad
  \textbf{Sujith Ravi\textsuperscript{1}} \quad
  \textbf{Dan Roth\textsuperscript{1}} \\[0.5em]
  \textsuperscript{1}Oracle AI Science \quad
  \textsuperscript{2}University of Pennsylvania
}

\begin{document}
\maketitle

\begin{abstract} 
Decoding strategies play a central role in shaping the reasoning ability of large language models (LLMs). Traditional methods such as greedy decoding and beam search often suffer from error propagation, while sampling-based approaches introduce randomness without adequate robustness. Self-consistency improves reliability by aggregating multiple rollouts, but incurs significant computational overhead.
We propose an entropy-guided decoding framework that introduces token-level adaptivity into generation. At each step, the model computes the entropy of the token distribution, identifies high-uncertainty positions, and selectively branches on these vulnerable points. A dynamic pool of partial rollouts is maintained and expanded until solutions are completed, concentrating computation where uncertainty is greatest and avoiding unnecessary exploration in confident regions. To enable efficient termination, we apply a rollout-level Entropy After </Think> (EAT) stopping criterion by performing entropy evaluation after the full reasoning trace, rather than incrementally at every step.
Experiments on GSM8K, AMC2023, and their perturbed variants demonstrate that our method achieves consistently strong accuracy. Notably, on smaller LLMs, performance is comparable to GPT-5 while operating at a fraction of the cost. 
\end{abstract}

\section{Introduction}
 
Large language models (LLMs) have achieved remarkable performance across a wide range of tasks, including natural language understanding, summarization, code generation and scientific question answering \citep{bommasani2021opportunities, chowdhery2023palm, openai2023gpt4}. Despite these successes, their reliability on complex reasoning tasks, such as mathematical problem solving, symbolic logic, and multi-step inference remains limited \cite{Wanga2022}. These tasks require maintaining long chains of precise intermediate steps, where a single early misstep can propagate unchecked and produce a final answer that appears logical and convincing yet is fundamentally incorrect \citep{boye2025llm_math_failures, zhang2025comprehension_without_competence}. In such settings, even a single erroneous token can alter the course of the reasoning process and irreversibly compromise the final outcome \citep{williamson2025syntactic}.

\begin{figure}
\centering
\includegraphics[width=8cm]{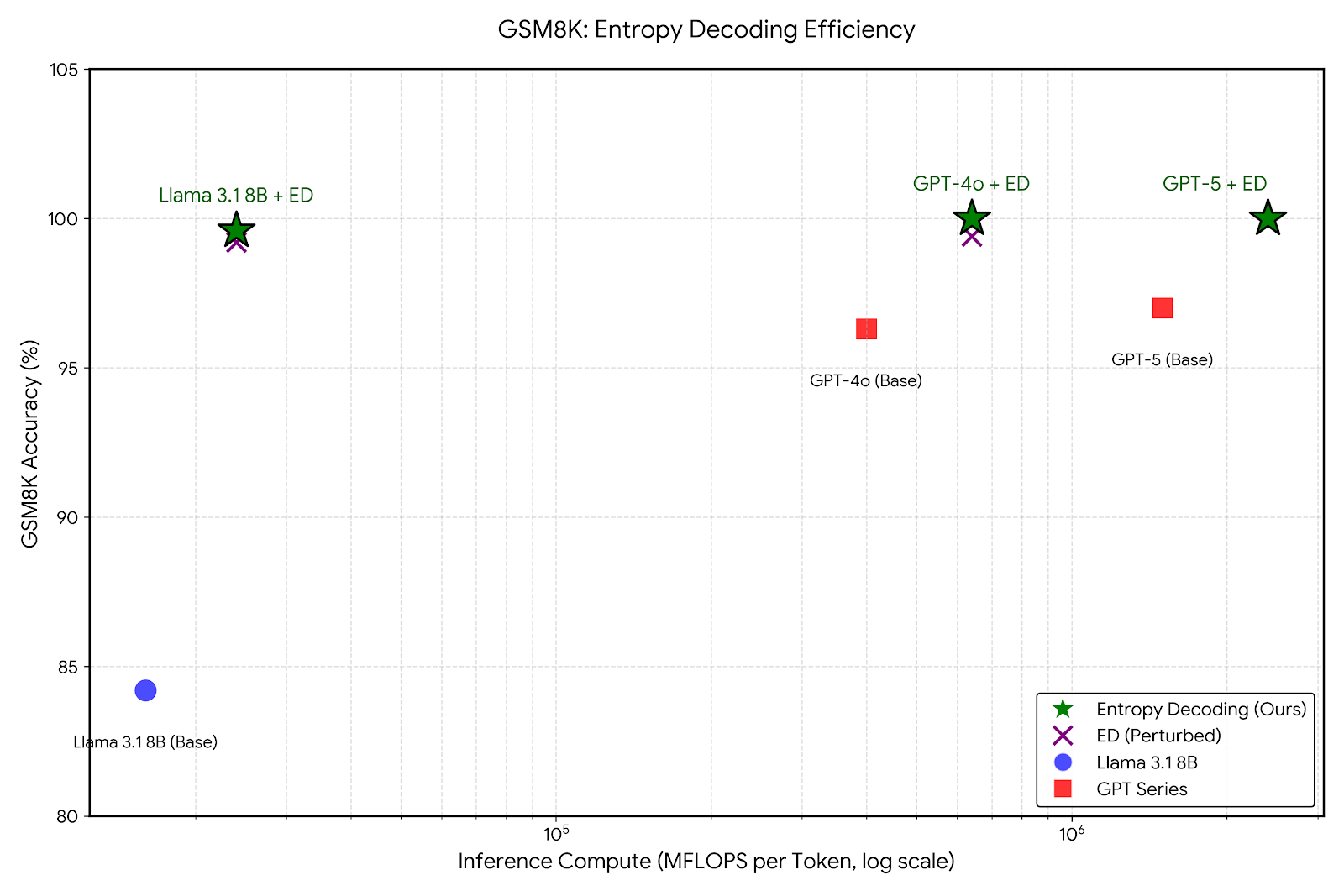}
\caption{Performance on GSM8K, our Entropy Decoding approach enables the 8B Llama model to outperform the base GPT models with $\sim$33x less compute.}
\vspace{-1.5em}
\label{fig:gsm8keff}
\end{figure}
 
A central challenge arises from the fact that standard decoding strategies, such as top-\(k\) sampling \cite{fan2018hierarchical, holtzman2019curious} and beam search \cite{sutskever2014sequence, wiseman2016sequence} are not designed to accommodate the brittleness of reasoning trajectories in large language models. Top-\(k\) sampling applies a fixed truncation at every decoding step, irrespective of whether the model is highly confident or deeply uncertain. This leads to inefficiencies: search effort is spent on unambiguous positions while decision-critical tokens receive insufficient exploration. Beam search, although maintains multiple hypotheses, expands candidates according to locally high probabilities. In reasoning tasks, such local bias can be misleading. An incorrect reasoning step may receive high probability due to patterns learned during pretraining, causing the beam to converge prematurely on flawed continuations while pruning the correct but lower probability trajectories \cite{yin2019benchmarking}. Consequently, both methods lack mechanisms to adapt their search behavior to the model's evolving uncertainty profile over time.

\begin{figure*}[ht]
\centering
\includegraphics[width=\linewidth]{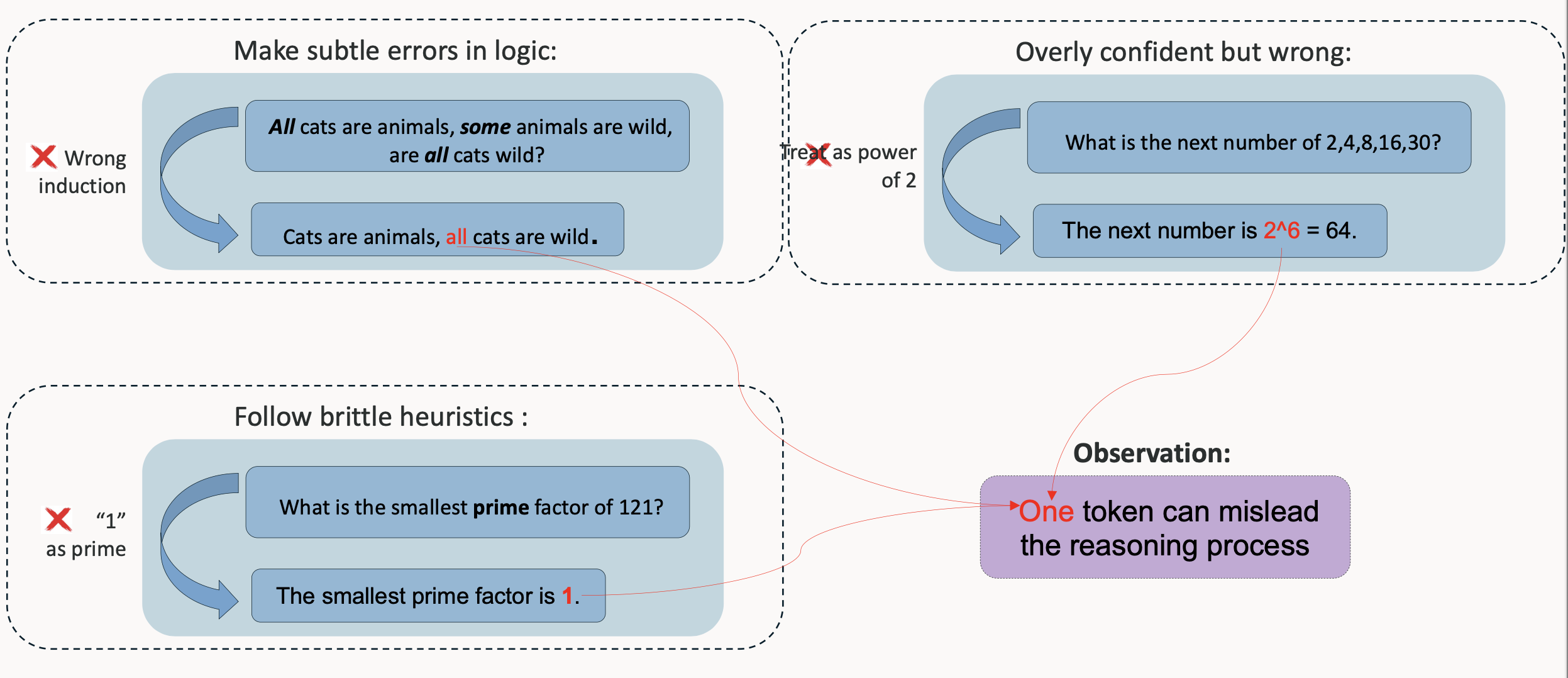}
\caption{Different kinds of reasoning errors made by LLMs in complex tasks are due to one or a few wrong tokens which mislead the problem-solving direction.}
\vspace{-1.5em}
\label{fig:reasoningerrors}
\end{figure*}

Our central observation is that decision-critical moments \cite{wang2025high} in reasoning often coincide with high-entropy tokens positions where the model's next-token probability distribution is relatively uniform. This uniformity reflects uncertainty about which reasoning step to take next, indicating that the model's internal knowledge is insufficient to decisively select one continuation. In contrast, low-entropy tokens, where a single choice dominates the distribution, offer limited potential gain from alternative exploration. Identifying and focusing search on high-entropy positions thus provides a principled way to allocate computational effort where it is most likely to affect correctness.

Based on this insight, we propose HN-decode, a high-entropy-token-guided decoding framework that allocates search budget adaptively rather than uniformly. HN-decode operates in two stages: (1) it generates an initial solution using a base decoding strategy; and (2) it scans the output to identify high-entropy positions, selectively expanding alternative continuations from these points while preserving low-entropy regions. This design mimics the behavior of a human problem solver who, upon encountering uncertainty, reconsiders multiple possible next steps before committing. By concentrating exploration on decision-critical moments, HN-decode improves robustness by reducing cascading reasoning errors, increases accuracy on challenging tasks, and achieves more efficient use of search resources than conventional decoding approaches.

Our contributions are threefold:

\begin{itemize}
    \item We identify the limitations of uniform decoding strategies in reasoning tasks and formalize the notion of decision-critical moments via token-wise entropy.

    \item We introduce HN-decode, a high-entropy-token-guided decoding/search framework that adaptively allocates search budget to uncertain positions, enabling targeted exploration without exhaustive expansion.

    \item We demonstrate that HN-decode improves solution accuracy and reasoning ability on challenging mathematical reasoning benchmarks across varying difficulty levels, while maintaining pareto search efficiency.
\end{itemize}

\section{Preliminaries}

\subsection{LLM Reasoning}

Large language models (LLMs), trained on massive and diverse text corpora, have shown surprising emergent reasoning capabilities, enabling them to solve tasks requiring multi-step inference, symbolic manipulation, and compositional problem solving. These capabilities are especially relevant in domains such as mathematical problem solving, scientific question answering, program synthesis, and formal logical reasoning, where arriving at a correct answer often demands a structured chain of intermediate steps rather than a single direct prediction \citep{wei2022chain, yang2025emergent, pan2023logiclm}.

Although different reasoning-oriented approaches improve accuracy and interpretability, they often share a fundamental limitation: the generation of intermediate reasoning steps is typically carried out in a single forward pass for each sampled trajectory. Errors made at early tokens can propagate irreversibly through subsequent steps, producing final answers that may appear coherent yet remain incorrect \cite{wei2022chain, zhang2024selfcorrection}. Figure \ref{fig:reasoningerrors} shows one such example.

Furthermore, most decoding strategies used in these methods apply uniform policies across all positions, without adaptively allocating search capacity to decision-critical steps where model uncertainty is highest. As a result, they do not alter the underlying left-to-right decoding dynamics at inference time; the model still commits to tokens sequentially without explicit mechanisms to explore or revise uncertain regions \cite{creswell2022faithful, welleck2022generating}. 

Consequently, reasoning-intensive tasks remain brittle, since a single incorrect inference step, such as a misapplied algebraic transformation, can invalidate the entire solution. Overcoming this brittleness requires inference-time methods that identify and revisit critical decision points rather than treating all tokens as equally reliable during generation.

\subsection{Decoding Strategies}

Large language models (LLMs) rely critically on decoding strategies to convert token-level probability distributions into coherent outputs. The choice of decoding method directly affects not only fluency and diversity, but also the correctness of reasoning-heavy tasks such as mathematics, programming, and scientific problem solving \citep{holtzman2020degeneration, zhu2024deductive}. Despite considerable progress, existing decoding approaches each exhibit limitations that constrain their effectiveness for complex reasoning.

Greedy decoding selects the highest-probability token at each step and is computationally efficient, but it is highly prone to error propagation: an early local mistake irreversibly derails the entire generation \citep{meister2020best}. Beam search extends greedy decoding by maintaining multiple candidate sequences, retaining the top-$k$ partial hypotheses. While this increases exploration, beam search often collapses to high-likelihood yet uninformative or repetitive trajectories \citep{meister2021determinantal}. Moreover, it optimizes likelihood rather than semantic or logical validity \citep{kasai2022clarity}, making it misaligned with reasoning objectives.

Stochastic methods such as top-$k$ sampling and nucleus (top-$p$) sampling \citep{holtzman2020degeneration} introduce randomness to mitigate determinism. Although these strategies improve diversity, they are agnostic to reasoning structure: a single mis-sampled token can derail multi-step reasoning chains, which is problematic in domains where correctness is fragile.

Self-consistency \citep{wang2023selfconsistency} has been proposed as a decoding-time strategy to improve reasoning robustness by sampling multiple reasoning trajectories and aggregating their final answers via majority voting. While effective, this approach is inefficient: many sampled trajectories are low-quality and consume computational resources without improving accuracy. Furthermore, self-consistency only evaluates end results and does not identify or correct intermediate reasoning errors.

\begin{figure*}[ht]
\centering
\includegraphics[width=\linewidth]{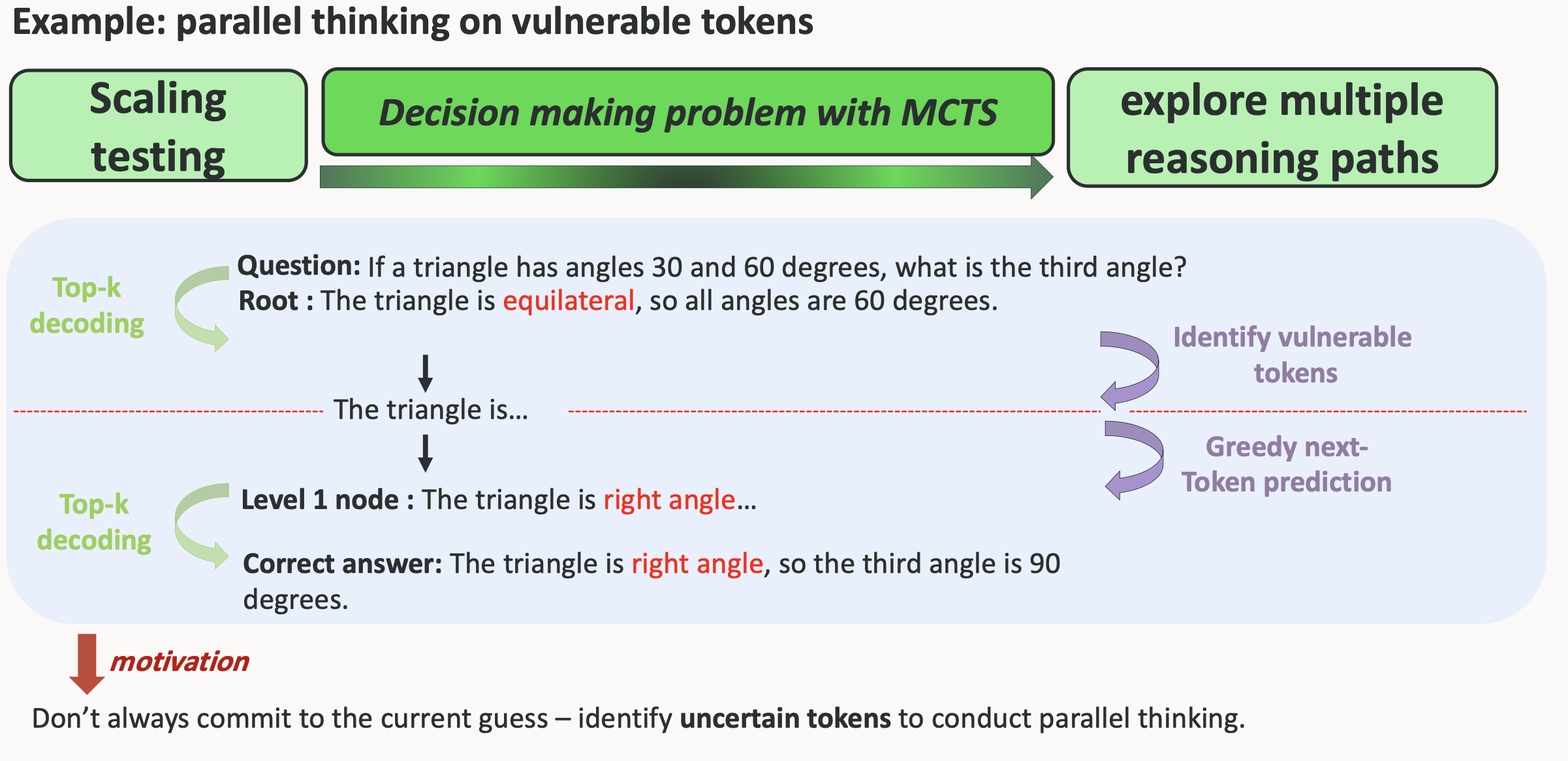}
\caption{An example of parallel thinking on vulnerable tokens. Here, "equilateral" is the vulnerable token.}
\vspace{-1.5em}
\label{fig:parallelThinking}
\end{figure*}

Across deterministic (greedy, beam) and stochastic (sampling, self-consistency) \cite{brown2020language, wang2023selfconsistency} strategies, a shared limitation emerges: all tokens are treated as equally reliable, with no adaptivity to local uncertainty. In practice, reasoning trajectories often hinge on vulnerable decision points (e.g. in Figure \ref{fig:parallelThinking}) where the model is most uncertain or misleading \citep{Shorinwa2025}. Existing decoding methods fail to identify or revisit these points, leading either to wasted computation or overconfidence in flawed continuations \cite{holtzman2019curious}. This gap motivates entropy-based decoding approaches \cite{qiu2025entropy,fei2025nudging,Xu25} that explicitly target uncertainty at the token level and guide exploration toward critical decision steps.

\section{Proposed Method}

We introduce an entropy-guided parallel decoding framework that departs from conventional left-to-right generation. Instead of committing to a single greedy rollout or exploring via indiscriminate sampling, our approach explicitly targets uncertain tokens points in the sequence where the model's distribution is high-entropy and prone to error. By branching at these tokens \cite{luo25} and maintaining a pool of partial rollouts, the method encourages parallel reasoning and allows multiple plausible continuations to be explored efficiently. Figure~\ref{fig:gsm8keff} shows our approach efficacy and key contribution, in that, we enable small LLMs to close the performance gap to top models' at a small fraction of the cost.

\subsection{Entropy-Based Token Selection}

At each decoding step, the model outputs a probability distribution over the vocabulary. We compute the Shannon entropy \cite{shannon} of this distribution to quantify uncertainty:

\begin{equation}
    H(p) = - \sum_i p_i \log p_i,
\end{equation} 

where $p\_i$ is the probability of the candidate token $i$. Tokens or positions with higher entropy indicate greater ambiguity, and are thus prime candidates for exploration. Rather than expanding every token equally, our algorithm allocates resources toward these vulnerable points, where errors are most likely to occur and corrections are most impactful.

\subsection{ Rollout Pool Construction}

Given a rollout generated using a standard decoding procedure (e.g. Top-k decoding), when a vulnerable token is identified, the algorithm does not random sample for more rollouts. Instead, it:

\begin{itemize}
    \item Branches: breaks the rollout at the top-k unconfident tokens to form a set of partial rollouts. Each partial rollouts contains tokens from the start of the rollout to the previous token of each high-entropy one.
    \item Stores: appends the corresponding partial rollouts into a job pool.
    \item  Continues: generation resumes by sampling a partial rollout from the pool, greedily decoding the next token, and then extending it to completion using the standard decoding procedure (e.g. Top-k decoding). 
\end{itemize}

This creates a dynamic set of reasoning paths, each diverging at critical points but reusing common low-entropy segments where the model is confident.

\subsection{Parallel Thinking via Partial Rollouts}

The job pool functions as a reservoir of alternative reasoning paths. Each time a candidate is sampled, greedy decoding is applied to generate the next token, in order to replace the vulnerable high-entropy one, it is rolled out to completion until a final answer is obtained. Figure~\ref{fig:parallelThinking} shows an example of parallel thinking on vulnerable tokens.  Correctness checks are then applied to the finished sequence (e.g., verifying the mathematical solution). Unlike self-consistency, which samples full rollouts indiscriminately, this approach concentrates computational resources on meaningful divergences, making exploration more targeted and efficient.

\subsection{Verification and Stopping Criterion}

Several prior methods \cite{EAT, graves2016adaptive, teerapittayanon2016branchynet, schuster2022confident, jazbec2024fast} employ confidence or uncertainty measures to enable early exiting during inference. The intuition is that a model's uncertainty, typically quantified as the entropy of its next-token distribution should decrease as reasoning progresses. A substantial drop in entropy (i.e., a high EAT score) after reasoning indicates that the model has likely reached a confident and stable conclusion.

Other lines of work have instead focused on training verifiers to assess the correctness of model generated solutions or to determine preference between candidate responses \cite{cobbe2021, Creswell2022, bai2022, zheng2023judging, Lightman2023, zhang2024small, paul2024refiner}. However, with recent advances in reasoning capabilities, LLMs can increasingly self-evaluate critiquing and refining their outputs without external verifiers often through natural language prompt templates that guide self-reflection and correction \cite{shinn2023reflexion, ling2023deductive, madaan2023self, weng2023large}.

In contrast, our approach leverages implicit correctness signals encoded within the model's own internal representations of the reasoning process. We employ a rollout-level EAT criterion, meaning entropy analysis is applied after the entire reasoning sequence (rather than step by step). Specifically, after generating a full reasoning rollout, we compute the entropy statistics namely, the mean and variance of next-token entropies at all </think> boundaries and use these to determine whether to accept, continue, or re-roll the reasoning sequence. A high EAT value suggests that the model has converged to a confident answer, whereas low EAT indicates lingering uncertainty or inconsistency, in which case additional rollouts may improve performance.

The procedure is as follows:

\begin{itemize}
\item Take a generated reasoning sequence.
\item Replay it token by token through the model.
\item Record next-token entropy at each </think> boundary.
\item Compute the mean ($\mu$) and variance ($\sigma$) of these entropies.
\item Accept the rollout if stability criteria are met (e.g., $\mu < \tau_1$ and $\sigma < \tau_2$), otherwise continue or re-roll. Here $\tau_1$ and $\tau_2$ are hyperparmeters which can be optimized on the training split of the datasets. 
\end{itemize}
   
\section{Experiments}
 
\begin{table*}[ht]
\centering
\begin{tabular}{lccccccccc}
\hline
\textbf{Model} &\textbf{Data} & \textbf{\makecell{Base\\ Accuracy}}  & \textbf{Accuracy} &  \textbf{\makecell{Mean \\ Jobs}} & \textbf{\makecell{Max\\ Jobs}} & \textbf{\makecell{Mean \\ Success \\ Job Rate}} & \textbf{\makecell{Mean \\ Success \\Depth}}& \textbf{\makecell{Mean \\ Elapsed \\ Time (s)}}   \\
\hline    
Llama3.1-8B & \texttt{GSM8k} & 84.2  &99.6 & 1.4 & 23 & 0.993  &0.997  &30.2   \\
Llama3.1-8B & \texttt{AMC}   &  45.0   & 97.5 &1.6&21 & 0.786 &1.21& 16.9     \\ 
GPT4o & \texttt{GSM8k} &  96.3  & 100 & 1.3 & 19 & 0.987  &0.996  &42.2   \\
GPT4o & \texttt{AMC}   & 75.0  & 98.25 & 1.9&12 & 0.759 &1.33& 51.0    \\
\hline
\end{tabular}
\caption{Results of entropy-guided decoding method on GSM8k and AMC2023 datasets. The table reports the base accuracy (\%): pass@1 of the base model,  accuracy (\%): accuracy of the proposed method, job statistics, mean success job rate, mean success depth and average elapsed time in seconds.}
\label{tab:results}
\end{table*}

We evaluate several large language models in this study: 
GPT-4o \citep{openai2024gpt4o}, and LLaMA-3.1-8B \citep{dubey2024llama}. Each model is tested under \textbf{few-shot Chain-of-Thought (CoT)} setting \citep{wei2022chain, Ton25}. 

The prompt is for all the aforementioned LLMs are the same, and it is (we also tested variations of the prompts and observed similar results):  
\begin{tcolorbox}[
  breakable,
  colback=gray!5,colframe=gray!60,
  boxrule=0.4pt,arc=2pt,
  left=6pt,right=6pt,top=4pt,bottom=4pt,
  listing only,
  listing options={
    basicstyle=\ttfamily\small,
    breaklines=true,
    columns=fullflexible,
    keepspaces=true,
    showstringspaces=false
  }
]
You are given a question between
the tags:
<|question|> and <|/question|>.\\[1pt]

<|question|>
\{user\_question\}
<|/question|>\\[1pt]

First, think about the question and provide a step-by-step reasoning process between the tags:\\[1pt]
<think>
...
</think>\\[1pt]

Finally, on a new line print only the final answer:
a single number with no extra text or formatting.
\end{tcolorbox}



% ...




\subsection{Setup}

We evaluate our entropy-guided decoding framework on two widely used reasoning benchmarks: GSM8k  (grade-school math word problems) \cite{GSM} and AMC2023 \cite{AMC2023} (American Mathematics Competition problems which are substantially harder). We used Llama3.1-8B and GPT4o for this task. We used the default hyparameters (top\_k, temperature).  The model is prompted with chain-of-thought reasoning, and decoding is performed with our method. We run this using the APIs \footnote{https://console.groq.com/docs/model/llama-3.1-8b-instant} and \footnote{https://platform.openai.com/docs/models/gpt-4o}. 
For each dataset, we record:
\begin{enumerate}
    \item Accuracy: proportion of tasks solved correctly.
    \item Base accuracy: proportion of tasks solved without branching.
    \item Job statistics: number of rollouts ("jobs") required before success.
    \item Success job rate: ratio of successful rollouts among total generated jobs. 
    \item Success depth: the depth at which branching first reaches a correct solution.
    \item Elapsed time: wall-clock latency per problem.
\end{enumerate}

We summarize the performance of our entropy-guided decoding method on GSM8k \cite{GSM} and AMC2023 \cite{AMC2023} datasets. Table \ref{tab:results} reports the accuracy, first-trial accuracy, job statistics, success job rate, and elapsed time statistics.

\subsection{GSM8k Results}

On GSM8K, using the Llama3.1-8B model, our method achieves 99.6\% accuracy, with 84.2\% of tasks solved correctly on the first trial. On average, only 1.4 attempts were required per problem, with a mean elapsed time of 30.2 seconds, and no corrupted or failed tasks were observed. These results demonstrate that our approach handles simpler arithmetic reasoning tasks extremely efficiently.

Using the GPT4o model, our method also yields a significant increase in accuracy, requiring only 42.2 seconds on average. This further confirms that our approach is efficient for simpler arithmetic reasoning tasks.

\subsection{AMC2023 Results}

On AMC2023, which is significantly more challenging, our method still achieves 97.5\% accuracy across 40 tasks for Llama3.1-8B and 98.25\% accuracy for GPT4o. However, only 45\% and 75\% of the problems were solved at the first trial for both models, requiring additional exploration of vulnerable tokens.  For Llama3.1-8B, the maximum number of jobs required for a single task was 21, with an average of 1.6 jobs created per task. For GPT4o, the maximum number of jobs required for a single task was 12, with an average of 1.9 jobs created per task.

Despite the increased complexity, our method successfully identified correct solutions for more tasks. Notably, correctness was consistently achieved within finite and bounded exploration budgets, demonstrating the scalability of entropy-guided rollouts. The variation in success job rate 
(\text{mean} $\approx 0.786$ for Llama3.1-8B, and \text{mean} $\approx 0.759$ for GPT4o) indicates that although some branches terminate unsuccessfully, the pool-based strategy ensures sufficient coverage of correct reasoning paths.

\subsection{Perturbed datasets and ablation studies}

\begin{table}[h]
\begin{tabular}{|p{6.5cm}|}
\hline
\textbf{Perturbed questions}  \\
\hline
    Betty is saving money for a new wallet which costs \$90. Betty has only forty percent of the money she needs. Her parents decided to give her \$10 for that purpose, and her grandparents two and a half times as much as her parents. How much more money does Betty need to buy the wallet?
      \\
\hline  
Julie is reading a 110-page book. Yesterday, she was able to read 5 pages and today, she read two and a half times as many pages as yesterday. If she wants to read 45\% of the remaining pages tomorrow, how many pages should she read? \\
\hline  
 Tina makes \$17.50 an hour. If she works more than 9 hours per shift, she is eligible for overtime, which is paid by your hourly wage + 0.4 times your hourly wage. If she works 9 hours every day for 4 days, how much money does she make? \\
 \hline  
 Randy has 55 mango trees on his farm. He also has 7 less than 0.35 times as many coconut trees as mango trees. How many trees does Randy have in all on his farm?  \\
\hline  
\end{tabular}
\caption{Examples of perturbed questions from the GSM8K.}
\label{tab:perturbation}
\end{table}
 
To assess model robustness on mathematical reasoning tasks, we generated larger evaluation sets following the perturbation method introduced in \cite{yu2024reasonagain}. These datasets contain modified versions of the original benchmarks and are designed to test whether models can reliably produce correct answers for different surface forms of the same underlying program. The approach in \cite{yu2024reasonagain} first extracts executable programs from the original math datasets and validates them using the corresponding input–output pairs, ensuring they capture the reasoning necessary to solve the text problems. GPT5 model then creates new questions by replacing the input–output pairs while preserving the program structure. We manually verify the correctness of te perturbed questions and provide the answers. Examples of the resulting perturbed dataset are shown in Table~\ref{tab:perturbation}. Using this method, we generated 2{,}242 perturbed questions from GSM8K and 320 from AMC2023.
We evaluates LLaMA-3.1-8B \citep{dubey2024llama} and GPT-4o \citep{openai2024gpt4o}.
Each model is tested using its default hyperparameters (e.g., top\_k, temperature) under the few-shot chain-of-thought (CoT) prompt setting \citep{wei2022chain}.
We apply our method to these perturbed datasets, with results summarized in Table~\ref{tab:resperturbed}. The findings indicate that standard LLMs struggle on the perturbed datasets, whereas our method achieves stable and robust performance.

\begin{table*}[ht]
\centering
\begin{tabular}{lcccccc}
\hline
\textbf{Model} &  \textbf{Data} &  \textbf{\makecell{Base\\ Accuracy}} & \textbf{\makecell{Proposed\\ Accuracy}}  & \textbf{\makecell{Random \\ Accuracy}} & \textbf{\makecell{Mean \\ Elapsed \\ Time(s) \\ (Proposed)}} & \textbf{\makecell{Mean \\ Elapsed \\ Time(s) \\ (Random) }}  \\
Llama3.1-8B & GSM8K\_p & 75.3 & 99.2 & 85.3 & 37.8& 95.4\\
Llama3.1-8B & AMC\_p & 42.7 &  98.3 & 92.7 & 17.4 &21.3 \\
GPT4o & GSM8K\_p & 79.5 & 99.4  & 94.5 & 45.7 & 96.9\\
GPT4o & AMC\_p & 43.2 & 91.9  & 87.6 & 57.4 & 93.8\\ 
\hline
\end{tabular}
\caption{Performance of LLMs on perturbed GSM8K and AMC2023 datasets. Base Accuracy refers to the percentage accuracy of the original model.
Proposed Accuracy is the accuracy (\%) achieved using our proposed method.
Random Accuracy is the accuracy (\%) obtained when regeneration is initiated from a randomly selected token.
Mean Elapsed Time (Proposed) is the average time per question when regeneration uses our proposed method.
Mean Elapsed Time (Random) is the average time per question when regeneration starts from a randomly chosen token.}
\vspace{-2em}
\label{tab:resperturbed}
\end{table*}

\subsection{Regeneration at Random Tokens}

For comparison, we also consider an alternative strategy in which a token is selected at random, and regeneration proceeds from that point using the same limits for consistency. The results of this approach are reported in the last column of Table~\ref{tab:resperturbed}.

\begin{figure}[ht]
\centering
\includegraphics[width=0.9\linewidth]{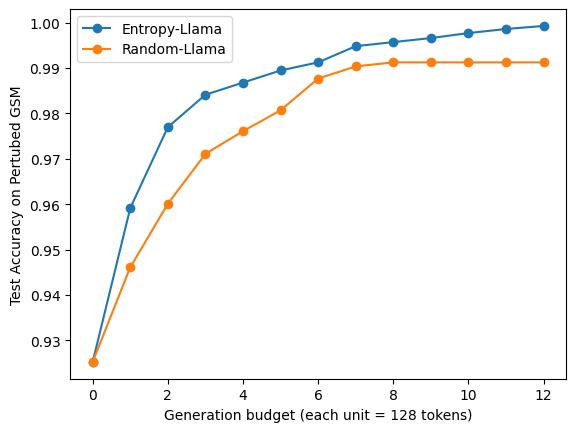}
\caption{  
Test accuracy on the perturbed GSM8K dataset under different generation budgets. Each x-axis unit represents 128 tokens. Results are shown for the Llama3.1-8B model using entropy-based and random approaches.}

\label{fig:accuracy_with_budget}
\end{figure}

We further evaluate performance across different generation budgets, defined as the maximum number of allowed generated tokens, to examine how much accuracy LLaMA-3.1-8B can achieve with both the proposed method and the random approach. These results are illustrated in Figure~\ref{fig:accuracy_with_budget}. 

Overall, selecting the highest-entropy token consistently outperforms random token selection under the same generation budget.

\subsection{Comparison with SOTA Model}

We compared the LLaMA-3.1-8B model, using our entropy-based decoding, against the state-of-the-art GPT-5 model in terms of both performance and cost on perturbed datasets.

For GPT-5, we used the hosted API\footnote{\url{https://platform.openai.com/docs/models/gpt-5}}, which charges \$10 per million output tokens and \$1.25 per million input tokens. Evaluation was performed on the perturbed GSM8K and AMC2023 datasets. The distributions of input tokens, output tokens, and inference times are detailed in ~\ref{app:costgpt5}.

For LLaMA-3.1-8B, we first assessed the cost via the hosted API\footnote{\url{https://console.groq.com/docs/model/llama-3.1-8b-instant}}, priced at \$0.08 per million output tokens and \$0.05 per million input tokens. Distributions of input tokens, output tokens, and runtime are shown in ~\ref{app:costllama}.

Since LLaMA-3.1-8B is open-source, we additionally ran it locally to evaluate on-premise inference costs. Full details of the setup and computations are provided in ~\ref{app:costlocalllama}.

Table~\ref{tab:gpt5vsllama} presents a direct comparison between GPT-5 and LLaMA-3.1-8B under our entropy-based decoding framework. Despite its smaller size, LLaMA-3.1-8B achieves comparable  accuracy with GPT-5, while operating at more than an order of magnitude lower cost.

\begin{table*}[ht]
\centering
\begin{tabular}{lcccccc}
\hline
\textbf{Model} &  \textbf{Data} &  \textbf{Accuracy} & \textbf{\makecell{Avg.  number \\ of input  tokens}}& \textbf{\makecell{Avg.  number \\ of output  tokens}} &  cost (\textcent)  \\
GPT5 & GSM8K\_p & 96.6  & 125 & 508 & 0.52 \\
Llama3.1-8B & GSM8K\_p& 99.2 & 3485 & 4323 & 0.05 \\ 
GPT5 & AMC\_p & 100  & 158 & 1779 & 1.8 \\
Llama3.1-8B &  AMC\_p &98.3 &  3692 & 5285 & 0.06 \\
\hline
\end{tabular}
\caption{We compare the GPT5 model, and the LLaMA-3.1-8B model under entropy-based decoding. Performance is evaluated on the perturbed GSM8K and AMC2023 datasets. Our analysis includes accuracy, the average number of input and output tokens, and the associated API cost.}
\label{tab:gpt5vsllama}
\end{table*}

\subsection{Hyperparameters and Ablation Studies}

The hyperparameters in our experiments were tuned using Optuna \cite{akiba2019optuna} to maximize accuracy on the GSM8K training split with LLaMA-3.1-8B. Although hyperparameters could be optimized individually for each model and dataset, we adopt a single configuration to highlight the effectiveness of our entropy-based decoding. The final hyperparameter values are reported in Table~\ref{tab:parameter}.

We also performed ablation studies by varying each hyperparameter, with results summarized in Table~\ref{tab:ablations} in ~\ref{appendix:ablation}. These results show that changes to the hyperparameters lead to only minor variations in model performance.




\subsection{Discussion}

The comparison between GSM8K and AMC2023, along with their perturbed versions, highlights the adaptive behavior of our framework. 
Unlike self-consistency, which generates a fixed number of full rollouts often resulting in redundant computation our approach dynamically allocates effort, branching only at uncertain tokens.
These findings demonstrate that entropy-guided decoding achieves both efficiency and robustness: it quickly solves simpler tasks while scaling gracefully to more challenging ones by concentrating exploration on the most uncertain points.

\section{Conclusions}

In this work, we introduced an entropy-guided decoding framework for LLMs, aimed at overcoming the limitations of conventional strategies such as greedy search, sampling, and self-consistency. Rather than committing prematurely to a single reasoning trajectory or generating multiple full rollouts indiscriminately, our method identifies high-entropy tokens points of uncertainty and selectively branches at these decision-critical positions. By maintaining a dynamic pool of partial rollouts, the framework enables parallel reasoning while allocating computational resources adaptively and efficiently. We also proposed a rollout-level EAT criterion, applying entropy analysis after the entire reasoning sequence, to decide whether to accept, continue, or re-roll the sequence.

Experimental results on the GSM8K and AMC2023 benchmarks, including their perturbed variants, demonstrate the robustness and effectiveness of our approach. 
These findings underscore the potential of uncertainty-aware decoding as a scalable and efficient approach to reasoning. By directly addressing error propagation at the token level, entropy-guided rollouts enable LLMs to solve tasks reliably and efficiently. Future work may extend this paradigm to broader domains such as code generation, scientific discovery, and multimodal reasoning, as well as explore integration with reinforcement learning signals to further refine process-level rewards.  
   
\section{Limitations}

Although entropy-guided decoding improves robustness and efficiency in reasoning tasks, several limitations remain. First, the method relies on entropy as a proxy for uncertainty, which may not always correlate with true reasoning difficulty. In cases where the model is confidently wrong, such as systematic misconceptions or spurious correlations, the framework may fail to trigger branching at critical steps. 
Second, while adaptive branching is more efficient than generating full rollouts, the approach can still incur high computational costs on problems with many consecutive uncertain tokens. The number of spawned jobs may grow rapidly in adversarial or highly combinatorial settings, especially when no correct trajectory is easily recoverable.
Third, the framework assumes access to a token-level entropy signal during inference, which may not be available or efficiently computed in all deployment environments. Integrating the method into API-restricted or latency-sensitive systems could therefore be challenging.
Fourth, hyperparameter selection could be optimized using the training splits of each dataset. However, in this work, we use a single hyperparameter setting to demonstrate the effectiveness of entropy-based decoding.
Finally, our evaluation focuses on mathematical reasoning benchmarks. Although the approach may generalize to domains such as code generation, scientific discovery, or multimodal reasoning, its effectiveness outside the tested datasets remains to be validated. Future work is needed to examine how domain structure, task format, and model scale affect the behavior of entropy-guided branching.

\bibliography{ref}

\begin{thebibliography}{55}
\providecommand{\natexlab}[1]{#1}

\bibitem[{Akiba et~al.(2019)Akiba, Sano, Yanase, Ohta, and Koyama}]{akiba2019optuna}
Takuya Akiba, Shotaro Sano, Toshihiko Yanase, Takeru Ohta, and Masanori Koyama. 2019.
\newblock {Optuna: A Next-Generation Hyperparameter Optimization Framework}.
\newblock In \emph{Proceedings of the 25th {ACM} {SIGKDD} International Conference on Knowledge Discovery \& Data Mining}, pages 2623--2631.

\bibitem[{Bai and et~al(2022)}]{bai2022}
Yuntao Bai and et~al. 2022.
\newblock Constitutional ai: Harmlessness from ai feedback.
\newblock \emph{https://arxiv.org/abs/2212.08073}.

\bibitem[{Bommasani et~al.(2021)Bommasani, Hudson, Adeli, and et~al.}]{bommasani2021opportunities}
Rishi Bommasani, Daniel Hudson, Ehsan Adeli, and et~al. 2021.
\newblock On the opportunities and risks of foundation models.
\newblock \emph{arXiv preprint arXiv:2108.07258}.

\bibitem[{Boye and Moell(2025)}]{boye2025llm_math_failures}
Johan Boye and Birger Moell. 2025.
\newblock Large language models and mathematical reasoning failures.
\newblock \emph{arXiv preprint arXiv:2502.11574}.

\bibitem[{Brown et~al.(2020)Brown, Mann, Ryder, and et~al.}]{brown2020language}
Tom~B. Brown, Benjamin Mann, Nick Ryder, and et~al. 2020.
\newblock Language models are few-shot learners.
\newblock \emph{arXiv preprint arXiv:2005.14165}.

\bibitem[{Chowdhery et~al.(2023)Chowdhery, Narang, Devlin, Bosma, Mishra, Roberts, Barham, Chung, Sutton, Gehrmann et~al.}]{chowdhery2023palm}
Aakanksha Chowdhery, Sharan Narang, Jacob Devlin, Maarten Bosma, Gaurav Mishra, Adam Roberts, Paul Barham, Hyung~Won Chung, Charles Sutton, Sebastian Gehrmann, and 1 others. 2023.
\newblock Palm: Scaling language modeling with pathways.
\newblock \emph{Journal of Machine Learning Research}, 24(240):1--113.

\bibitem[{Cobbe and et~al(2021)}]{cobbe2021}
Karl Cobbe and et~al. 2021.
\newblock \href {https://arxiv.org/abs/2110.14168} {Training verifiers to solve math word problems}.
\newblock \emph{arXiv preprint arXiv:2509.26522}.

\bibitem[{Creswell and Shanahan(2022{\natexlab{a}})}]{creswell2022faithful}
Antonia Creswell and Murray Shanahan. 2022{\natexlab{a}}.
\newblock \href {https://arxiv.org/abs/2208.14271} {Faithful reasoning using large language models}.
\newblock \emph{arXiv preprint arXiv:2208.14271}.

\bibitem[{Creswell and Shanahan(2022{\natexlab{b}})}]{Creswell2022}
Antonia Creswell and Murray Shanahan. 2022{\natexlab{b}}.
\newblock Faithful reasoning using large language models.
\newblock \emph{https://arxiv.org/abs/2208.14271}.

\bibitem[{et~al.(2022)}]{wei2022chain}
Wei et~al. 2022.
\newblock Chain-of-thought prompting elicits reasoning in large language models.
\newblock In \emph{NeurIPS}.

\bibitem[{Fan et~al.(2018)Fan, Lewis, and Dauphin}]{fan2018hierarchical}
Angela Fan, Mike Lewis, and Yann Dauphin. 2018.
\newblock Hierarchical neural story generation.
\newblock In \emph{ACL}.

\bibitem[{Fei et~al.(2025)Fei, Razeghi, and Singh}]{fei2025nudging}
Yu~Fei, Yasaman Razeghi, and Sameer Singh. 2025.
\newblock Nudging: Inference-time alignment of llms via guided decoding.
\newblock In \emph{Proceedings of the 63rd Annual Meeting of the Association for Computational Linguistics (Volume 1: Long Papers)}, pages 12702--12739.

\bibitem[{Graves(2016)}]{graves2016adaptive}
Alex Graves. 2016.
\newblock Adaptive computation time for recurrent neural networks.
\newblock \emph{arXiv preprint arXiv:1603.08983}.

\bibitem[{Holtzman et~al.(2020{\natexlab{a}})Holtzman, Buys, Du, Forbes, and Choi}]{holtzman2019curious}
Ari Holtzman, Jan Buys, Li~Du, Maxwell Forbes, and Yejin Choi. 2020{\natexlab{a}}.
\newblock The curious case of neural text degeneration.
\newblock In \emph{ICLR}.

\bibitem[{Holtzman et~al.(2020{\natexlab{b}})Holtzman, Buys, Du, Forbes, and Choi}]{holtzman2020degeneration}
Ari Holtzman, Jan Buys, Li~Du, Maxwell Forbes, and Yejin Choi. 2020{\natexlab{b}}.
\newblock \href {https://openreview.net/forum?id=rygGQyrFvH} {The curious case of neural text degeneration}.
\newblock In \emph{International Conference on Learning Representations}.

\bibitem[{Jazbec et~al.(2024)Jazbec, Timans, Veljkovic, Sakmann, Zhang, N{\ae}sseth, and Nalisnick}]{jazbec2024fast}
Metod Jazbec, Alexander Timans, Tin~H Veljkovic, Kaspar Sakmann, Dan Zhang, Christian~A N{\ae}sseth, and Eric Nalisnick. 2024.
\newblock Fast yet safe: Early-exiting with risk control.
\newblock In \emph{Advances in Neural Information Processing Systems}, volume~37, pages 129825--129854.

\bibitem[{Kaggle(2023)}]{AMC2023}
Kaggle. 2023.
\newblock \href {https://www.kaggle.com/competitions/amc2023} {Amc - regression final competition}.

\bibitem[{Kasai et~al.(2022)Kasai, Sakaguchi, Bras, Radev, Choi, and Smith}]{kasai2022clarity}
Jungo Kasai, Keisuke Sakaguchi, Ronan~Le Bras, Dragomir Radev, Yejin Choi, and Noah~A. Smith. 2022.
\newblock \href {https://arxiv.org/abs/2204.05424} {A call for clarity in beam search: How it works and when it stops}.
\newblock \emph{arXiv preprint arXiv:2204.05424}.

\bibitem[{Lightman and et~al(2023)}]{Lightman2023}
Hunter Lightman and et~al. 2023.
\newblock Let’s verify step by step.
\newblock \emph{https://arxiv.org/abs/2305.20050}.

\bibitem[{Ling et~al.(2023)Ling, Fang, Li, Huang, Lee, Memisevic, and Su}]{ling2023deductive}
Zhan Ling, Yunhao Fang, Xuanlin Li, Zhiao Huang, Mingu Lee, Roland Memisevic, and Hao Su. 2023.
\newblock Deductive verification of chain-of-thought reasoning.
\newblock \emph{Advances in Neural Information Processing Systems}, 36:36407--36433.

\bibitem[{Llama~Team(2024)}]{dubey2024llama}
Meta Llama~Team, AI. 2024.
\newblock \href {https://arxiv.org/pdf/2407.21783} {The llama 3 herd of models}.

\bibitem[{Luo et~al.(2025)Luo, Fang, Liu, and Bai}]{luo25}
GuangSheng Luo, ZhiJun Fang, JianLing Liu, and YiFanBai Bai. 2025.
\newblock Clip guided image caption decoding based on monte carlo tree search: Clip guided image caption decoding.
\newblock \emph{Multimedia Systems}.

\bibitem[{Madaan et~al.(2023)Madaan, Tandon, Gupta, Hallinan, Gao, Wiegreffe, Alon, Dziri, Prabhumoye, Yang et~al.}]{madaan2023self}
Aman Madaan, Niket Tandon, Prakhar Gupta, Skyler Hallinan, Luyu Gao, Sarah Wiegreffe, Uri Alon, Nouha Dziri, Shrimai Prabhumoye, Yiming Yang, and 1 others. 2023.
\newblock Self-refine: Iterative refinement with self-feedback.
\newblock \emph{Advances in Neural Information Processing Systems}, 36:46534--46594.

\bibitem[{Meister et~al.(2021)Meister, Forster, and Cotterell}]{meister2021determinantal}
Clara Meister, Michael Forster, and Ryan Cotterell. 2021.
\newblock \href {https://aclanthology.org/2021.acl-long.512.pdf} {Determinantal beam search}.
\newblock In \emph{Proceedings of ACL}.

\bibitem[{Meister et~al.(2020)Meister, Vieira, and Cotterell}]{meister2020best}
Clara Meister, Tim Vieira, and Ryan Cotterell. 2020.
\newblock \href {https://aclanthology.org/2020.acl-main.563.pdf} {Best-first beam search}.
\newblock In \emph{Proceedings of ACL}.

\bibitem[{{OpenAI}(2021)}]{GSM}
{OpenAI}. 2021.
\newblock Grade school math (gsm8k) dataset.
\newblock \url{https://github.com/openai/grade-school-math}.

\bibitem[{OpenAI(2023)}]{openai2023gpt4}
OpenAI. 2023.
\newblock {GPT-4} technical report.
\newblock In \emph{arXiv preprint arXiv:2303.08774}.

\bibitem[{OpenAI(2024)}]{openai2024gpt4o}
OpenAI. 2024.
\newblock \href {https://arxiv.org/abs/2410.21276} {Gpt-4o system card}.

\bibitem[{Pan et~al.(2023)Pan, Albalak, Wang, and Wang}]{pan2023logiclm}
Liangming Pan, Alon Albalak, Xinyi Wang, and William~Yang Wang. 2023.
\newblock \href {https://arxiv.org/abs/2305.12295} {Logic-lm: Empowering large language models with symbolic solvers for faithful logical reasoning}.
\newblock \emph{arXiv preprint arXiv:2305.12295}.

\bibitem[{Paul et~al.(2024)Paul, Ismayilzada, Peyrard, Borges, Bosselut, West, and Faltings}]{paul2024refiner}
Debjit Paul, Mete Ismayilzada, Maxime Peyrard, Beatriz Borges, Antoine Bosselut, Robert West, and Boi Faltings. 2024.
\newblock Refiner: Reasoning feedback on intermediate representations.
\newblock In \emph{Proceedings of the 18th Conference of the European Chapter of the Association for Computational Linguistics (Volume 1: Long Papers)}, pages 1100--1126.

\bibitem[{Qiu et~al.(2025)Qiu, Ou, Wu, Li, Liu, and King}]{qiu2025entropy}
Zexuan Qiu, Zijing Ou, Bin Wu, Jingjing Li, Aiwei Liu, and Irwin King. 2025.
\newblock Entropy-based decoding for retrieval-augmented large language models.
\newblock In \emph{Proceedings of the 2025 Conference of the Nations of the Americas Chapter of the Association for Computational Linguistics: Human Language Technologies (Volume 1: Long Papers)}, pages 4616--4627.

\bibitem[{Schuster et~al.(2022)Schuster, Fisch, Gupta, Dehghani, Bahri, Tran, Tay, and Metzler}]{schuster2022confident}
Tal Schuster, Adam Fisch, Jai Gupta, Mostafa Dehghani, Dara Bahri, Vinh Tran, Yi~Tay, and Donald Metzler. 2022.
\newblock Confident adaptive language modeling.
\newblock In \emph{Advances in Neural Information Processing Systems}, volume~35, pages 17456--17472.

\bibitem[{Shannon(1948)}]{shannon}
Claude~E. Shannon. 1948.
\newblock A mathematical theory of communication.
\newblock \emph{Bell System Technical Journal}, 27(3):379--423.

\bibitem[{Shinn et~al.(2023)Shinn, Cassano, Gopinath, Narasimhan, and Yao}]{shinn2023reflexion}
Noah Shinn, Federico Cassano, Ashwin Gopinath, Karthik Narasimhan, and Shunyu Yao. 2023.
\newblock Reflexion: Language agents with verbal reinforcement learning.
\newblock \emph{Advances in Neural Information Processing Systems}, 36:8634--8652.

\bibitem[{Shorinwa et~al.(2025)Shorinwa, Mei, Lidard, Ren, and MajumdarAuthors}]{Shorinwa2025}
Ola Shorinwa, Zhiting Mei, Justin Lidard, Allen~Z. Ren, and Anirudha MajumdarAuthors. 2025.
\newblock A survey on uncertainty quantification of large language models: Taxonomy, open research challenges, and future directions.
\newblock \emph{ACM Computing Surveys}.

\bibitem[{Sutskever et~al.(2014)Sutskever, Vinyals, and Le}]{sutskever2014sequence}
Ilya Sutskever, Oriol Vinyals, and Quoc~V Le. 2014.
\newblock Sequence to sequence learning with neural networks.
\newblock In \emph{NeurIPS}.

\bibitem[{Teerapittayanon et~al.(2016)Teerapittayanon, McDanel, and Kung}]{teerapittayanon2016branchynet}
Surat Teerapittayanon, Bradley McDanel, and Hsiang-Tsung Kung. 2016.
\newblock Branchynet: Fast inference via early exiting from deep neural networks.
\newblock In \emph{2016 23rd International Conference on Pattern Recognition (ICPR)}, pages 2464--2469. IEEE.

\bibitem[{Ton et~al.(2025)Ton, Taufiq, and Liu}]{Ton25}
Jean-Francois Ton, Muhammad~Faaiz Taufiq, and Yang Liu. 2025.
\newblock Understanding chain-of-thought in llms through information theory.
\newblock \emph{ICML}.

\bibitem[{Wang and et~al(2022)}]{wang2025high}
Shenzhi Wang and et~al. 2022.
\newblock \href {https://arxiv.org/abs/2506.01939} {Beyond the 80/20 rule: High-entropy minority tokens drive effective reinforcement learning for llm reasoning}.
\newblock \emph{arXiv preprint arXiv:2506.01939}.

\bibitem[{Wang et~al.(2025)Wang, McInerney, Wang, and Kallus}]{EAT}
Xi~Wang, James McInerney, Lequn Wang, and Nathan Kallus. 2025.
\newblock \href {https://arxiv.org/pdf/2509.26522} {Eat: Entropy after </think> for reasoning model early exiting}.
\newblock \emph{arXiv preprint arXiv:2509.26522}.

\bibitem[{Wang et~al.(2023)Wang, Wei, Schuurmans, Le, Chi, Narang, Chowdhery, and Zhou}]{wang2023selfconsistency}
Xuezhi Wang, Jason Wei, Dale Schuurmans, Quoc Le, Ed~Chi, Sharan Narang, Aakanksha Chowdhery, and Denny Zhou. 2023.
\newblock \href {https://arxiv.org/abs/2203.11171} {Self-consistency improves chain of thought reasoning in language models}.
\newblock \emph{arXiv preprint arXiv:2203.11171}.

\bibitem[{Wang et~al.(2024)Wang, Zhang, Li, Eisenschlos, Perot, Wang, Miculicich, Fujii, Shang, Lee, and Pfister}]{zhang2024selfcorrection}
Zilong Wang, Hao Zhang, Chun-Liang Li, Julian~Martin Eisenschlos, Vincent Perot, Zifeng Wang, Lesly Miculicich, Yasuhisa Fujii, Jingbo Shang, Chen-Yu Lee, and Tomas Pfister. 2024.
\newblock \href {https://arxiv.org/abs/2401.04398} {Chain-of-table: Evolving tables in the reasoning chain for table understanding}.
\newblock \emph{arXiv preprint arXiv:2401.04398}.

\bibitem[{Wanga et~al.(2022)Wanga, Wanga, Shena, Wua, Zhoua, and Chandraa}]{Wanga2022}
Ruonan Wanga, Runxi Wanga, Yunwen Shena, Chengfeng Wua, Qinglin Zhoua, and Rohitash Chandraa. 2022.
\newblock Evaluation of llms for mathematical problem solving.
\newblock \emph{https://arxiv.org/pdf/2506.00309}.

\bibitem[{Welleck et~al.(2022)Welleck, Brantley, West, and Choi}]{welleck2022generating}
Sean Welleck, Keith Brantley, Peter West, and Yejin Choi. 2022.
\newblock \href {https://aclanthology.org/2022.emnlp-main.38.pdf} {Generating sequences with repeated sampling}.
\newblock In \emph{Proceedings of the 2022 Conference on Empirical Methods in Natural Language Processing}.

\bibitem[{Weng et~al.(2023)Weng, Zhu, Xia, Li, He, Liu, Sun, Liu, and Zhao}]{weng2023large}
Yixuan Weng, Minjun Zhu, Fei Xia, Bin Li, Shizhu He, Shengping Liu, Bin Sun, Kang Liu, and Jun Zhao. 2023.
\newblock Large language models are better reasoners with self-verification.
\newblock In \emph{Findings of the Association for Computational Linguistics: EMNLP 2023}, pages 2550--2575.

\bibitem[{Williamson et~al.(2025)Williamson, Ji, and Dwyer}]{williamson2025syntactic}
Dane~A Williamson, Yangfeng Ji, and Matthew~B Dwyer. 2025.
\newblock Syntactic blind spots: How misalignment leads to llms’ mathematical errors.
\newblock In \emph{Proceedings of The 3rd Workshop on Mathematical Natural Language Processing (MathNLP 2025)}, pages 1--14.

\bibitem[{Wiseman and Rush(2016)}]{wiseman2016sequence}
Sam Wiseman and Alexander~M Rush. 2016.
\newblock Sequence-to-sequence learning as beam-search optimization.
\newblock In \emph{EMNLP}.

\bibitem[{Xu et~al.(2025)Xu, Zhang, and Liu}]{Xu25}
Jinxuan Xu, Yuqian Zhang, and Hang Liu. 2025.
\newblock A survey of uncertainty estimation methods on large language models.
\newblock \emph{https://arxiv.org/abs/2503.00172}.

\bibitem[{Yang et~al.(2025)Yang, Campbell, Huang, Wang, Cohen, and Webb}]{yang2025emergent}
Yukang Yang, Declan~Iain Campbell, Kaixuan Huang, Mengdi Wang, Jonathan~D. Cohen, and Taylor~Whittington Webb. 2025.
\newblock \href {https://openreview.net/forum?id=y1SnRPDWx4} {Emergent symbolic mechanisms support abstract reasoning in large language models}.
\newblock \emph{OpenReview / ICML}.

\bibitem[{Yin et~al.(2019)Yin, Hay, and Roth}]{yin2019benchmarking}
Wenpeng Yin, Jessica Hay, and Dan Roth. 2019.
\newblock Benchmarking zero-shot text classification: Datasets, evaluation and entailment approach.
\newblock In \emph{EMNLP}.

\bibitem[{Yu et~al.(2024)Yu, Zhou, Cheng, and Roth}]{yu2024reasonagain}
Xiaodong Yu, Ben Zhou, Hao Cheng, and Dan Roth. 2024.
\newblock Reasonagain: Using extractable symbolic programs to evaluate mathematical reasoning.
\newblock \emph{arXiv preprint arXiv:2410.19056}.

\bibitem[{Zhang et~al.(2024)Zhang, Khalifa, Logeswaran, Kim, Lee, Lee, and Wang}]{zhang2024small}
Yunxiang Zhang, Muhammad Khalifa, Lajanugen Logeswaran, Jaekyeom Kim, Moontae Lee, Honglak Lee, and Lu~Wang. 2024.
\newblock Small language models need strong verifiers to self-correct reasoning.
\newblock In \emph{Findings of the Association for Computational Linguistics ACL 2024}, pages 15637--15653.

\bibitem[{Zhang(2025)}]{zhang2025comprehension_without_competence}
Zheng Zhang. 2025.
\newblock Comprehension without competence: Architectural limits of llms in symbolic computation and reasoning.
\newblock \emph{arXiv preprint arXiv:2507.10624}.

\bibitem[{Zheng et~al.(2023)Zheng, Chiang, Sheng, Zhuang, Wu, Zhuang, Lin, Li, Li, Xing et~al.}]{zheng2023judging}
Lianmin Zheng, Wei-Lin Chiang, Ying Sheng, Siyuan Zhuang, Zhanghao Wu, Yonghao Zhuang, Zi~Lin, Zhuohan Li, Dacheng Li, Eric Xing, and 1 others. 2023.
\newblock Judging llm-as-a-judge with mt-bench and chatbot arena.
\newblock \emph{Advances in neural information processing systems}, 36:46595--46623.

\bibitem[{Zhu et~al.(2024)Zhu, Zhang, Xie, and Su}]{zhu2024deductive}
Tinghui Zhu, Kai Zhang, Jian Xie, and Yu~Su. 2024.
\newblock \href {https://arxiv.org/abs/2401.17686} {Deductive beam search: Decoding deducible rationale for chain-of-thought reasoning}.
\newblock \emph{arXiv preprint arXiv:2401.17686}.

\end{thebibliography}

\clearpage
\appendix

\section{Appendix}

\subsection{Example where the base model (llama3.1-8B) fails but the proposed method succeeds}

Table~\ref{tab:examplesuccess} presents a case in which the base model (Llama3.1-8B) fails, but the proposed method succeeds. Additional examples where the base model fails but our method works are provided in Table~\ref{tab:successeamples}.

\begin{table*}[h] 
\centering
\begin{tabular}{|p{2.2cm}|p{4.2cm}|p{4.2cm}|} 
\hline
\textbf{Perturbed questions} &  \textbf{Base}  & \textbf{Proposed} \\
\hline 
 Issac has to buy pens and pencils for the upcoming school year. Issac buys 108 writing utensils total. He buys 12 more than 5 times the number of pencils than pens. How many pens did Issac buy? & 
Let's define the variables:

- Let \( x \) be the number of pens Issac buys.

- Let \( y \) be the number of pencils Issac buys.

According to the problem, he buys a total of 108 writing utensils:

\[ x + y = 108 \]

It is also given that he buys "12 more than 5 times the number of pencils than pens". This translates to:
\[
\textcolor{red}{x} = 5y + 12
\]

Now we have a system of two equations:

1. \( x + y = 108 \)

2. \( x = 5y + 12 \)

Substitute the expression for \( x \) from equation 2 into equation 1:
\[ 5y + 12 + y = 108 \]

Combine the \( y \) terms:
\[ 6y + 12 = 108 \]

Subtract 12 from both sides:
\[ 6y = 96 \]

Divide by 6:
\[ y = 16 \]

Now that we know the number of pencils, we can find the number of pens:
Substitute \( y = 16 \) back into equation 2:
\[ x = 5(16) + 12 \]
\[ x = 80 + 12 \]
\[ x = 92 \]

Issac buys 92 pens.  &
Let's define the variables:

- Let \( x \) be the number of pens Issac buys.

- Let \( y \) be the number of pencils Issac buys.

According to the problem, he buys a total of 108 writing utensils:

\[ x + y = 108 \]

It is also given that he buys "12 more than 5 times the number of pencils than pens". This translates to:

\[ y = 5x + 12 \]

Now we have a system of two equations:

1. \( x + y = 108 \)

2. \( y = 5x + 12 \)

Substitute the expression for \( y \) from equation 2 into equation 1:

\[ 5x + 12 + x = 108 \]

Combine the \( x \) terms:

\[ 6x + 12 = 108 \]

Subtract 12 from both sides:
\[ 6x = 96 \]

Divide by 6:
\[ x = 16 \]
 
Issac buys 16 pens.  
  
\\
\hline  
\end{tabular}
\caption{GSM8K example where the base model (llama3.1-8B) fails but the proposed method succeeds. \textcolor{red}{x} is an high entropy token.}  
\label{tab:examplesuccess}
\end{table*}

\begin{table*}[h] 
\centering
\begin{tabular}{|p{10.2cm}|p{1.2cm}|p{1.5cm}|} 
\hline
\textbf{Perturbed questions} &  \textbf{Base} & \textbf{Proposed} \\
\hline
Carlos is planting a lemon tree. The tree will cost \$90 to plant. Each year it will grow 7 lemons, which he can sell for \$1.5 each. It costs $\$3$ a year to water and feed the tree. How many years will it take before he starts earning
money on the lemon tree? & 12 & 13 \\
\hline  
Christina is planning a birthday party and needs .75 gift bags per invited guest, because 1/4 of attendees don't show up. She invited 16 friends. Gift bags are \$2 each. How much will she spend?&18 & 24 \\
\hline  
 Lee used to be able to run the 400-meter hurdles two seconds faster than Gerald would run the 400-meter hurdles.  But Gerald changed his diet, which improved his speed by 10\%.  If Lee runs the 400-meter hurdles in 38 seconds, how fast can Gerald, with his improved diet, run the 400-meter hurdles, in seconds? &32.4 &36 \\
 \hline  
 Adrien's total salary was 30 percent higher than Lylah's. Four years later, his salary had increased, and he was earning 40\% more than what he was making four years ago. If Adrien's and Lylah's salary increased simultaneously, and Adrien earned \$40000 four years ago, calculate the total salary the two were receiving four years later? &96000& 95200 \\
 \hline  
 A garden is filled with 105 flowers of various colors. There are twice as many red flowers as orange. There are five fewer yellow flowers than red. If there are 10 orange flowers, how many pink and purple flowers are there if they have the same amount and there are no other colors? &60 &30 \\
  \hline  
 Albert wants a paintbrush that costs \$1.50, a set of paints that costs \$4.35, and a wooden easel that costs \$12.65. Albert already has \$6.50. How much more money does Albert need? & 13 & 12 \\
 \hline  
 A banana tree has 100 bananas left after Raj cut some bananas from it. If Raj has eaten 70 bananas and has twice as many remaining in his basket, how many bananas were on the tree initially?  &240 & 310\\
 \hline  
 Drew is reseeding his lawn with grass seed. One bag of grass seed covers 250 square feet of lawn. His lawn is 22 feet from the house to the curb and 36 feet from side to side. He bought four bags of seed. How many extra square feet could the leftover grass seed cover after Drew reseeds his lawn? & 0 & 208 \\
 \hline
 Issac has to buy pens and pencils for the upcoming school year. Issac buys 108 writing utensils total. He buys 12 more than 5 times the number of pencils than pens. How many pens did Issac buy? &92 & 16 \\
\hline  
\end{tabular}
\caption{GSM8K examples where the base model fails but the proposed method succeeds.}
\label{tab:successeamples}
\end{table*}

\subsection{Time, input and output token distribution of the GPT5 model}
\label{app:costgpt5}

\begin{figure}[h]
\centering
\includegraphics[width=\linewidth]{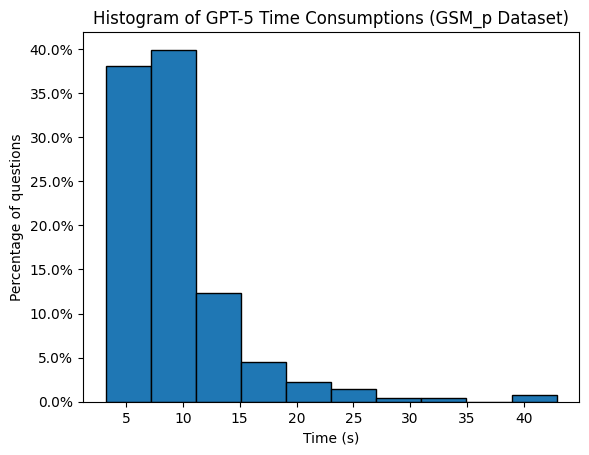}
\caption{\label{fig:timegpt5_gsmp8k}  
Consumption time (s) distribution of GPT5 model on the perturbed GSM8K dataset.}
\end{figure}

\begin{figure}[ht]
\centering
\includegraphics[width=\linewidth]{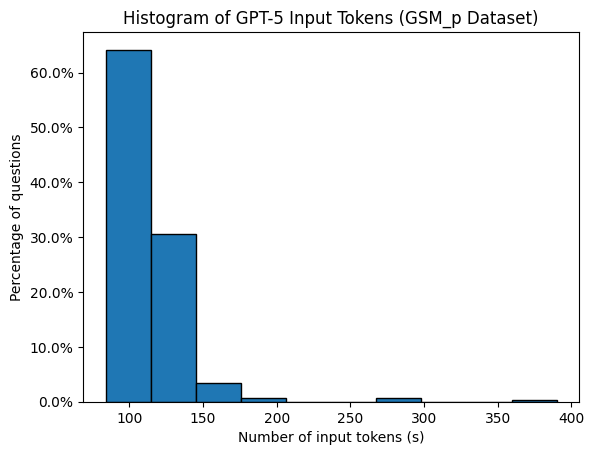}
\caption{\label{fig:inputtokens_gpt5_gsmp8k}  
Distribution of the number of input tokens for the GPT5 model on the perturbed GSM8K dataset.}
\end{figure}

\begin{figure}[ht]
\centering
\includegraphics[width=\linewidth]{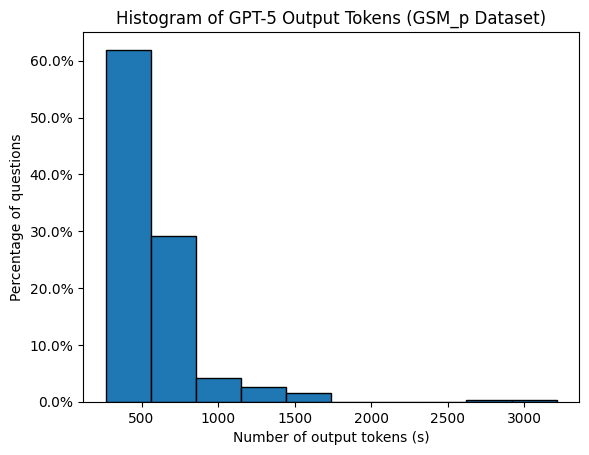}
\caption{\label{fig:outputtokens_gpt5_gsmp8k}  
Distribution of the number of output tokens for the GPT5 model on the perturbed GSM8K dataset.}
\end{figure}

\begin{figure}[h]
\centering
\includegraphics[width=\linewidth]{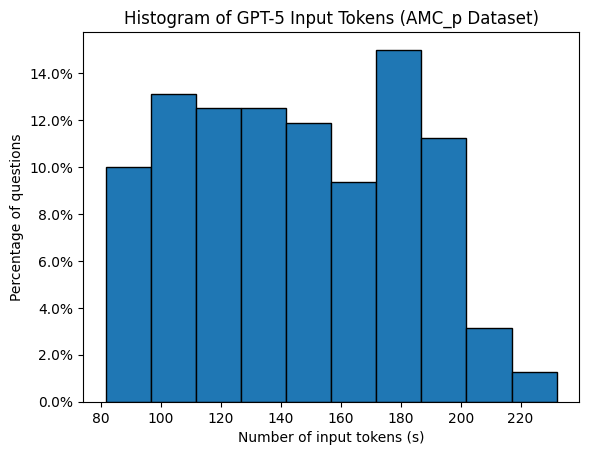}
\caption{\label{fig:inputtokens_gpt5_amc}  
Distribution of the number of input tokens for the GPT5 model on the perturbed AMC dataset.}
\end{figure}

\begin{figure}[h]
\centering
\includegraphics[width=\linewidth]{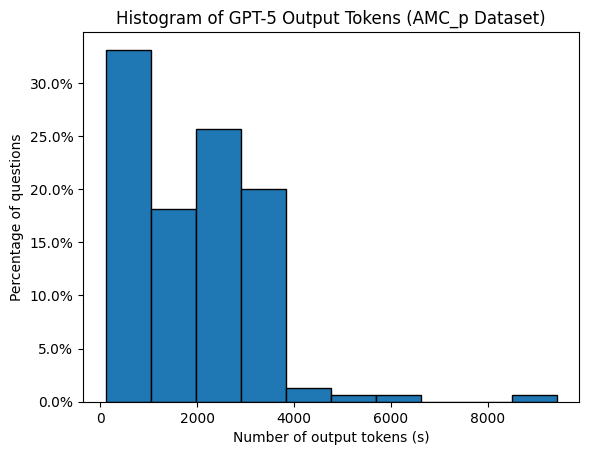}
\caption{\label{fig:outputtokens_gpt5_amc}  
Distribution of the number of output tokens for the GPT5 model on the perturbed AMC dataset.}
\end{figure}

\begin{figure}[h]
\centering
\includegraphics[width=\linewidth]{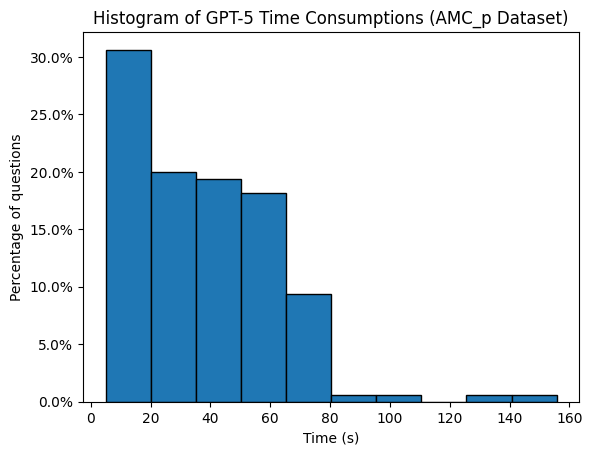}
\caption{\label{fig:timegpt5_gpt_amc}  
Consumption time (s) distribution of GPT5 model on the perturbed AMC dataset.}
\end{figure}

We present the distributions of consumed time, input tokens, and output tokens using the GPT5 model. The results for the perturbed GSM and AMC datasets are shown in Figure~\ref{fig:timegpt5_gsmp8k}, \ref{fig:inputtokens_gpt5_amc}, \ref{fig:outputtokens_gpt5_amc}, \ref{fig:timegpt5_gpt_amc}, \ref{fig:inputtokens_gpt5_gsmp8k}, and \ref{fig:outputtokens_gpt5_gsmp8k}.

\subsection{Time, input and output token distribution of the Llama3.1-8B model}
\label{app:costllama}

We present the distributions of consumed time, input tokens, and output tokens using the Llama3.1-8B model. The results for the perturbed GSM and AMC datasets are shown in Figure~\ref{fig:timellama_gsmp8k}, \ref{fig:inputtokens_llama_gsmp8k}, and \ref{fig:outputtokens_llama_gsmp8k}.

\begin{figure}[ht]
\centering
\includegraphics[width=\linewidth]{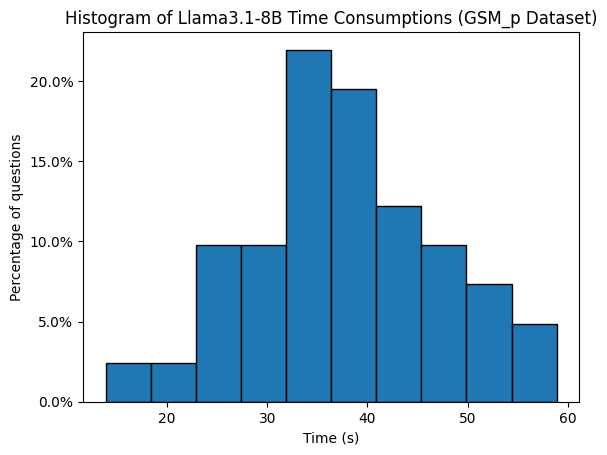}
\caption{\label{fig:timellama_gsmp8k}  
Consumption time (s) distribution of Llama3.1-8B model on the perturbed GSM8K dataset.}
\end{figure}

\begin{figure}[ht]
\centering
\includegraphics[width=\linewidth]{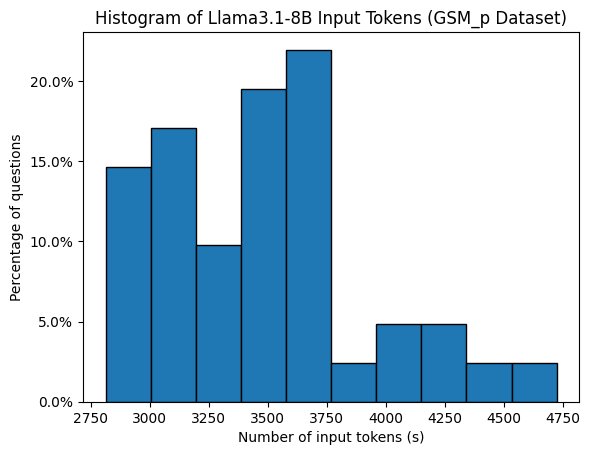}
\caption{\label{fig:inputtokens_llama_gsmp8k}  
Distribution of the number of input tokens for the Llama3.1-8B model on the perturbed GSM8K dataset.}
\end{figure}

\begin{figure}[ht]
\centering
\includegraphics[width=\linewidth]{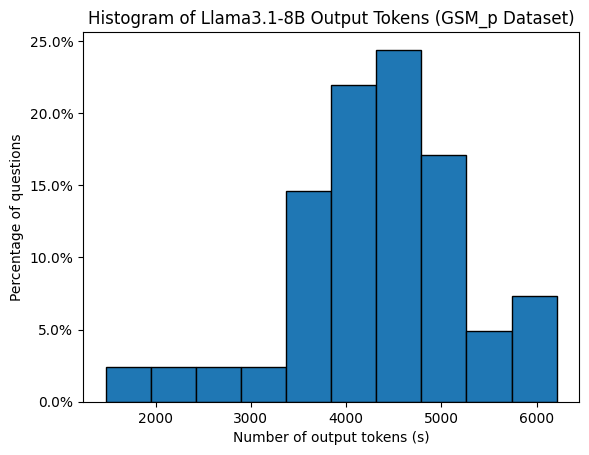}
\caption{\label{fig:outputtokens_llama_gsmp8k}  
Distribution of the number of output tokens for the Llama3.1-8B model on the perturbed GSM8K dataset.}
\end{figure}

\begin{figure}[ht]
\centering
\includegraphics[width=\linewidth]{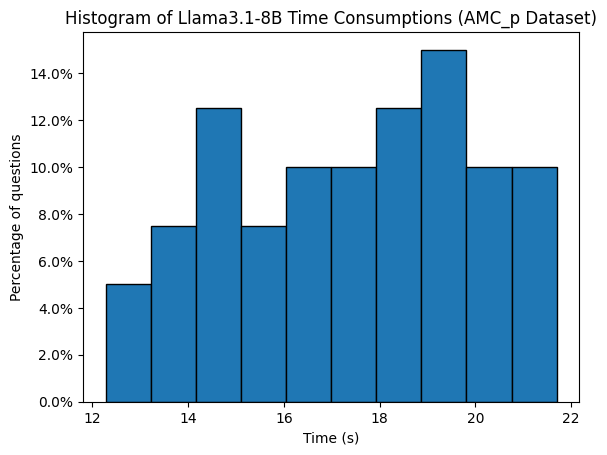}
\caption{\label{fig:timellama_gsmp8k}  
Consumption time (s) distribution of Llama3.1-8B model on the perturbed AMC dataset.}
\end{figure}

\begin{figure}[ht]
\centering
\includegraphics[width=\linewidth]{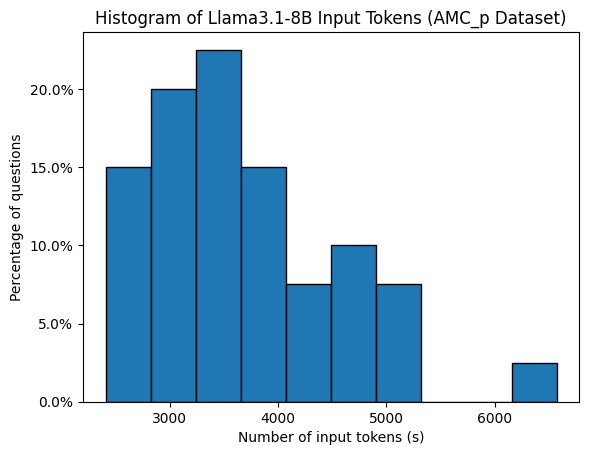}
\caption{\label{fig:inputtokens_llama_gsmp8k}  
Distribution of the number of input tokens for the Llama3.1-8B model on the perturbed AMC dataset.}
\end{figure}

\begin{figure}[ht]
\centering
\includegraphics[width=\linewidth]{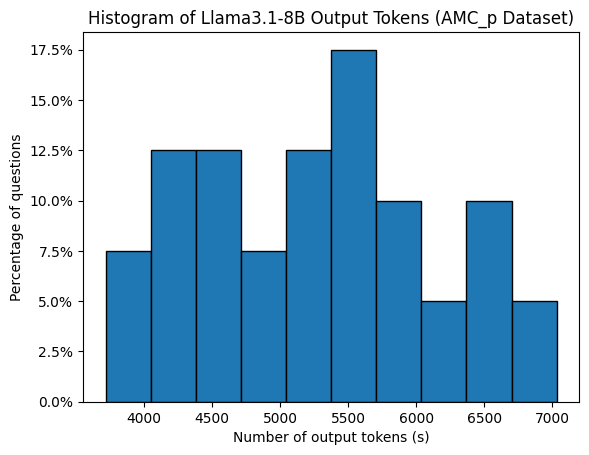}
\caption{\label{fig:outputtokens_llama_gsmp8k}  
Distribution of the number of output tokens for the Llama3.1-8B model on the perturbed AMC dataset.}
\end{figure}

\subsection{Parameter values}
\label{appendix:parameters}
We optimized the hyperparameters using Optuna \cite{akiba2019optuna}, targeting the highest accuracy on the GSM8K training split using Llama3.1-8B. While dataset and model specific tuning is possible, we intentionally use one shared set of hyperparameters to demonstrate that entropy-based decoding remains effective without per-dataset optimization. The chosen values are reported in Table~\ref{tab:parameter}.

\begin{table*}[h]
\centering
\begin{tabular}{|l|p{9.6cm}|c|}
\hline
\textbf{parameter} & \textbf{explanation} & \textbf{value} \\
\hline   
\texttt{max\_degree}      & The maximum number of entropy-heavy tokens to branch from.  & 3     \\ 
\texttt{min\_degree}      & The minimum number of entropy-heavy tokens to branch from.  & 2    \\ 
\texttt{degree\_depth\_decay} & How quickly branching tapers off as moving deeper. & 0.6 \\
\texttt{max\_mcts\_depth} & How deep any branch can go. & 3 \\
\texttt{max\_num\_create\_jobs} & How many total branches can be explored. & 32 \\ 
\texttt{$\tau_1$} & The mean entropy threshold for stopping & 2.3 \\
\texttt{$\tau_2$} & The variance entropy threshold for stopping & 9.8 \\
\hline
\end{tabular}
\caption{Hyperparameters used in our experiments, along with explanations and the values chosen. These values were optimized to maximize accuracy on the GSM8K train split.}
\label{tab:parameter}
\end{table*}

\subsection{Ablation studies on parameter values}
\label{appendix:ablation}
We independently varied each hyperparameter to analyze its effect on both accuracy and computational efficiency. The results are shown in Table \ref{tab:ablations}.

\begin{table*}[ht]
\centering
\begin{tabular}{lccccc}
\hline
\textbf{Parameter} &  \textbf{Value} &  \textbf{Data}&  \textbf{Accuracy}& \textbf{\makecell{Mean  Elapsed  Time(s)}}  \\
\texttt{max\_degree} & 2 &  GSM8K\_p &   98.1 &  35.2  \\
\texttt{max\_degree} &  3 &  GSM8K\_p &   99.2 &  37.8 \\
\texttt{max\_degree} & 4 &  GSM8K\_p &   98.7 &  36.3  \\
\texttt{max\_degree} & 5 &  GSM8K\_p &  98.3 &  37.1  \\

\texttt{max\_degree} & 2 &  AMC\_p &   97.5 &  18.1  \\
\texttt{max\_degree} &  3 &  AMC\_p &   98.3 &  21.3\\
\texttt{max\_degree} & 4 &  AMC\_p &   99.1 &  22.7  \\
\texttt{max\_degree} & 5 &  AMC\_p &  99.8 &  28.2 \\

\texttt{min\_degree} & 1 &  GSM8K\_p &   98.5 &  32.9 \\
\texttt{min\_degree} &  2 &  GSM8K\_p &   99.2 &  37.8 \\

\texttt{min\_degree} & 1 &  AMC\_p &   97.8 &  19.7 \\
\texttt{min\_degree} &  2 &  AMC\_p &  98.3 &  21.3 \\

\texttt{degree\_depth\_decay} &  0.4 &  GSM8K\_p &   98.1 &  36.6 \\ 
\texttt{degree\_depth\_decay} &  0.5 &  GSM8K\_p &   99.0 &  39.2 \\ 
\texttt{degree\_depth\_decay} &  0.6 &  GSM8K\_p &   99.2 &  37.8 \\ 
\texttt{degree\_depth\_decay} &  0.7 &  GSM8K\_p &   96.9 &  35.7 \\

\texttt{degree\_depth\_decay} &  0.4 &  AMC\_p &   99.2 &  26.7 \\ 
\texttt{degree\_depth\_decay} &  0.5 &  AMC\_p &   98.6 &  22.9 \\ 
\texttt{degree\_depth\_decay} &  0.6 &  AMC\_p &  98.3 &  21.3 \\ 
\texttt{degree\_depth\_decay} &  0.7 &  AMC\_p &   97.8 &  19.9 \\

\texttt{max\_mcts\_depth} &  2 &  GSM8K\_p &   93.5 &  32.4 \\
\texttt{max\_mcts\_depth} &  3 &  GSM8K\_p &   99.2 &  37.8 \\
\texttt{max\_mcts\_depthy} &  4 &  GSM8K\_p &   99.4 &  39.7 \\

\texttt{max\_mcts\_depth} &  2 &  AMC\_p &   97.9 &  20.2 \\
\texttt{max\_mcts\_depth} &  3 &  AMC\_p &   98.3 &  21.3 \\
\texttt{max\_mcts\_depthy} &  4 &  AMC\_p &   99.0 &  27.9 \\

\texttt{max\_num\_create\_jobs} &  16 &  GSM8K\_p &   90.9 &  26.1 \\
\texttt{max\_num\_create\_jobs} &  32 &  GSM8K\_p &   99.2 &  37.8 \\
\texttt{max\_num\_create\_jobs} &  64 &  GSM8K\_p &   99.8 &  40.1 \\

\texttt{max\_num\_create\_jobs} &  16 &  AMC\_p &   96.3 &  18.6 \\
\texttt{max\_num\_create\_jobs} &  32 &  AMC\_p &   98.3 &  21.3 \\
\texttt{max\_num\_create\_jobs} &  64 &  AMC\_p &   99.2 &  27.9 \\

\texttt{$\tau_1$} & 2.0 &  GSM8K\_p &   98.0 &  36.3 \\
\texttt{$\tau_1$} & 2.3 &  GSM8K\_p &   99.2 &  37.8 \\
\texttt{$\tau_1$} & 3.0 &  GSM8K\_p &   98.8 &  38.1 \\

\texttt{$\tau_1$} & 2.0 &  AMC\_p &   98.8 &  27.1.3 \\
\texttt{$\tau_1$} & 2.3 &  AMC\_p &   98.3 &  21.3 \\
\texttt{$\tau_1$} & 3.0 &  AMC\_p &   97.3 &  18.8 \\

\texttt{$\tau_2$} & 9 &  GSM8K\_p &   98.6 &  39.1 \\
\texttt{$\tau_2$} & 9.8 &  GSM8K\_p &   99.2 &  37.8 \\
\texttt{$\tau_2$} & 11 &  GSM8K\_p &   97.9 &  36.6 \\

\texttt{$\tau_2$} & 9  & AMC\_p &   99.2 &  25.3 \\
\texttt{$\tau_2$} & 9.8   & AMC\_p &   98.3 &  21.3  \\
\texttt{$\tau_2$} & 11 &   AMC\_p &   97.5 &  19.2 \\
 
\hline
\end{tabular}
\caption{Ablation studies on different parameter values. Performance on the perturbed GSM8K and AMC2023 dataset using Llama3.1-8B with entropy decoding.   
Mean Elapsed Time is the average time (s) per question when regeneration uses our proposed method. }
\label{tab:ablations}
\end{table*}

\subsection{Cost analysis of running LLaMA-3.1-8B locally}
\label{app:costlocalllama}

Because LLaMA-3.1-8B is open-source, we additionally deploy it locally to evaluate inference cost.
The model is hosted on \textbf{8×A100 GPUs (40GB each)}. To estimate cost, we reference cloud GPU pricing from Lambda Labs\footnote{https://lambda.ai/pricing}
, which lists an hourly rate of \textbf{\$1.29 per A100-40GB}.
Running locally reduces end-to-end latency by approximately \textbf{90\%} compared to the API. Using the average runtime per query, we multiply compute time by GPU cost to obtain per-question estimates. The resulting cost is lower than API usage—approximately \textbf{0.03\textcent} per perturbed GSM8K question and \textbf{0.04\textcent} per perturbed AMC2023 question.

\end{document}